\documentclass{article}
\usepackage{icomp2026_conference,times}
\icompfinalcopy
\usepackage{amsmath}
\usepackage{amssymb}
\usepackage{amsfonts}
\usepackage{graphicx}
\usepackage{booktabs}
\usepackage{bbold}
\usepackage{amsthm}
\usepackage{float}
\usepackage{placeins}
\usepackage{algorithm}
\usepackage{algpseudocode}

\usepackage{amsmath,amsfonts,bm}

\def\eqref#1{equation~\ref{#1}}

\def\1{\bm{1}}

\DeclareMathAlphabet{\mathsfit}{\encodingdefault}{\sfdefault}{m}{sl}
\SetMathAlphabet{\mathsfit}{bold}{\encodingdefault}{\sfdefault}{bx}{n}

\newcommand{\CreditCompression}{2.8\%}
\newcommand{\CreditDeltaUtility}{-0.001}
\newcommand{\NsgaMeanHvChange}{+0.016}

\usepackage{hyperref}
\usepackage{url}
\usepackage{cleveref}

\title{Min-Marginal Power Indices for \\ Multi-Objective Feature Selection}

\author{Wenjie Shi \& Chi Zhao\thanks{Corresponding author.} \\
Saint Petersburg State University \\
7/9 Universitetskaya nab., \\
Saint Petersburg, 199034, Russia \\
\texttt{st147487@student.spbu.ru}, \texttt{chi.zhao@spbu.ru}}

\begin{document}
\maketitle
\lhead{}
\renewcommand{\headrulewidth}{0pt}

\begin{abstract}

Game-theoretic feature selection scores each feature by its average marginal contribution across coalitions. An average shows whether a feature helps overall, but not whether, in some coalition, adding it turns an acceptable model into an unacceptable one. Building on minimum marginal aggregation, we define the Min-Marginal Power Index (MinPI) on a binary utility-threshold game. For a fixed quota, MinPI takes the value $-1$, $0$, or $+1$ and labels each feature harmful, neutral, or decisive. A harmful label carries its own evidence, an evaluated coalition in which adding the feature drops utility below the quota, so no score cutoff is needed; finite-budget estimators keep this property but may miss harmful features. We use MinPI to propose feature removals, which are accepted only if they pass a validation gate, and to prefilter multi-objective search. On six benchmark datasets over five seeds, MinPI removed 12.5\% of analyzed features with almost no loss of held-out utility. On Madelon and Abalone, where it removed the most, its removals cost far less held-out utility than random removals of the same size. Prefiltering increased the dataset-averaged hypervolume of multi-objective search in all five seeds. MinPI is therefore a conservative feature-deletion step whose gains are largest where the data contain many removable features.
\end{abstract}

\section{Introduction}
\label{sec:intro}

Feature selection balances predictive utility against the number of retained variables and is naturally multi-objective~\citep{guyon2003introduction,chandrashekar2014survey,xue2016survey}. Filters rank features independently of the final model, wrappers search over subsets with the learner itself~\citep{kohavi1997wrappers}, and embedded methods such as recursive feature elimination discard features using the fitted model's weights~\citep{guyon2002gene}. Cooperative-game methods treat features as players and the utility of a model refitted on each coalition as the coalition value~\citep{cohen2007feature}; the Shapley value and the Banzhaf index summarize average marginal contributions~\citep{shapley1953value,banzhaf1965weighted}, and related selectors include Powershap~\citep{verhaeghe2022powershap} and ShapG~\citep{zhao2025shapg}. SHAP~\citep{lundberg2017unified} and its tree-ensemble form TreeSHAP~\citep{lundberg2020local} instead attribute the predictions of a single fitted model; removal-based explanations place these scores in one framework~\citep{covert2021explaining}, and Banzhaf indices have been used to prune the features of a fixed classifier without changing its decisions~\citep{kulynych2017feature}.

Removing a feature can help when it is harmful in some context: filtering harmful features can improve single-task models~\citep{lippl2024inductive}. A mean marginal contribution answers whether a feature helps on average, not whether, in some coalition, adding it turns an acceptable model into an unacceptable one, and Shapley-value-based scores have known limitations as feature-importance measures~\citep{kumar2020problems}. MinShap~\citep{zheng2026minshap} replaces averaging with minimum aggregation, but its continuous score still needs a separate cutoff; we specialize the same aggregation to a binary utility-threshold game, which defines the Min-Marginal Power Index (MinPI), yielding the values $-1$, $0$, and $+1$, where every $-1$ comes with an evaluated coalition in which adding the feature drops utility below the quota, so no second score cutoff is needed.

This paper makes four contributions. First, we define MinPI on a utility-threshold game: for a fixed quota, its value $-1$, $0$, or $+1$ labels a feature as harmful, neutral, or decisive, and every $-1$ is certified by an evaluated coalition, so no score cutoff is needed. Second, we separate exact computation from finite-budget estimation: observed harmful labels are verified by a found coalition, while missed witnesses remain possible. Third, we use $\alpha$ to generate compression--utility operating points and pass candidate removals through a validation-only gate. Fourth, we report a five-seed analysis on six benchmark datasets, showing that the method's strongest results arise on datasets with abundant removable structure. The analysis addresses four research questions: whether validation-gated removal of MinPI's harmful candidates preserves held-out utility, compared with score-based and random deletion of the same size (RQ1); how the removal operating point changes with the quota $\alpha$ (RQ2); whether MinPI prefiltering changes the outcome of multi-objective search under a fixed evaluation budget (RQ3); and how sensitive the recovered harmful labels are to the estimator, its guidance, and the utility (RQ4).

\section{Method}
\label{sec:method}

\subsection{Utility-Threshold Voting Game}

Let $N=\{1,\ldots,n\}$ denote the analyzed features. For $S\subseteq N$, let $\mathcal{M}_S$ be the fixed model family fitted using $S$, and let $U(\mathcal{M}_S)$ be a configured real-valued, higher-is-better utility. The empty coalition uses a constant predictor derived only from the training response: class priors for classification and the training mean for regression. With full-feature utility $U_{\mathrm{full}}=U(\mathcal{M}_N)>0$ and quota $\alpha\in(0,1]$, define
\begin{equation}
v_\alpha(S)=\mathbb{1}\!\left[U(\mathcal{M}_S)\geq \alpha U_{\mathrm{full}}\right].
\label{eq:utvg}
\end{equation}
Unlike a classical monotone simple game~(\citealp{owen1972multilinear}; \citealp{straffin1988shapley}), $v_\alpha$ need not be monotone: adding a feature can push utility just below the quota, and this single threshold flip is exactly the event MinPI certifies. Its marginal is
$\Delta_i(S)=v_\alpha(S\cup\{i\})-v_\alpha(S)\in\{-1,0,+1\}$. A coalition $S\subseteq N\setminus\{i\}$ with $\Delta_i(S)=-1$ is a \emph{harmful witness} for feature $i$.

\subsection{Min-Marginal Power Indices}

For a permutation $\pi$ of $N$, let $S_\pi^i$ be the features preceding $i$. We define the minimum Shapley--Shubik (MinSS) and minimum Banzhaf (MinBz) marginals
\begin{align}
\operatorname{MinSS}_i &= \min_{\pi}\Delta_i(S_\pi^i), \\
\operatorname{MinBz}_i &= \min_{S\subseteq N\setminus\{i\}}\Delta_i(S).
\end{align}

For every feature, $\operatorname{MinSS}_i=\operatorname{MinBz}_i\in\{-1,0,+1\}$, because each marginal lies in $\{-1,0,+1\}$ and every coalition $S\subseteq N\setminus\{i\}$ is the predecessor set $S_\pi^i$ of at least one permutation, so both minima range over the same marginals. The value is $-1$ exactly when some coalition is winning before $i$ is added and losing afterward; it is $+1$ exactly when every coalition is losing before $i$ is added and winning afterward; otherwise it is $0$. The value $-1$ is self-certifying: an evaluated coalition witnesses the flip from winning to losing, so the harmful label requires no additional score threshold. We call these classes \emph{harmful}, \emph{decisive}, and \emph{neutral at the worst-case level}, respectively, always relative to the fixed model family, data split, utility, and quota. The last term means neither harmful nor universally decisive in the exact game; it does not mean that the feature has zero predictive value.

The equality of the minima does not assert equality of the standard weighted Shapley--Shubik~\citep{shapley1954method} and Banzhaf~\citep{banzhaf1965weighted} indices. We write $\operatorname{MinPI}_i$ for either exact minimum analogue.

\subsection{Finite-Budget Estimation}

Exact evaluation requires at most $2^n$ distinct coalition utilities and $O(n2^{n-1})$ marginal comparisons. It is practical only at small $n$. For each feature, Algorithm A first samples a coalition size uniformly from $\{0,\ldots,n-1\}$ and then samples a subset of that size uniformly without replacement from $N\setminus\{i\}$, in the permutation-sampling spirit of Shapley-effects estimation~\citep{song2016shapley}. For a budget of $K\geq1$ sampled coalitions per feature, it returns the minimum observed marginal over the frozen schedule $\mathcal{S}_K$ of these coalitions:
\begin{equation}
\widehat{\operatorname{MinPI}}_i
=\min_{S\in\mathcal{S}_K}\Delta_i(S)
\geq \min_{S\subseteq N\setminus\{i\}}\Delta_i(S).
\label{eq:sample-min}
\end{equation}
The inequality holds because $\mathcal{S}_K\subseteq 2^{N\setminus\{i\}}$. Thus an observed $-1$ certifies a harmful witness, whereas a sampled $0$ or $+1$ does not exclude an unseen smaller marginal.

Algorithm B uses training-only mutual information (MI) and absolute correlation---along the lines of ShapG's correlation-graph guidance~\citep{zhao2025shapg}---to construct a seed coalition $B_i$: up to ten correlation neighbors are included and then padded from the mutual-information ranking toward $\min(20,\lfloor n/3\rfloor)$. We refer to this construction as MI-correlation guidance. In each trial, every seed member is removed independently with probability $.3$; an addition count is drawn uniformly from $\{0,\ldots,\min(5,n-1)\}$ and that many currently absent features are taken without replacement from a random order. The resulting support is restricted by this construction: Algorithm B proposes only coalitions with at most $\min(5,n-1)$ features outside $B_i$, so witnesses farther from $B_i$ are never proposed, and even when every coalition is reachable, as on eight-feature Abalone, its mass is concentrated near $B_i$. Let $P_i$ denote the distribution of one perturbation for feature $i$. If a harmful witness $S_i^*$ has mass $\rho_i=P_i(S_i^*)>0$ and the $K_{\mathrm{adv}}\geq1$ draws are independent, then, since each draw misses $S_i^*$ with probability $1-\rho_i$,
\begin{equation}
\Pr[\text{detect a harmful witness}]\geq 1-(1-\rho_i)^{K_{\mathrm{adv}}}.
\label{eq:detection}
\end{equation}
 Algorithm A, in contrast, draws every coalition $S\subseteq N\setminus\{i\}$ with probability $1/[n\binom{n-1}{|S|}]>0$, so~\eqref{eq:detection} holds for it with $K$ draws and $\rho_i\geq1/[n\binom{n-1}{\lfloor(n-1)/2\rfloor}]$. To make the support probabilities in Figure~\ref{fig:overview}(a,b) auditable, consider $N=\{f_0,\ldots,f_4\}$ and the analytically specified game
\begin{equation}
v(S)=\mathbb{1}\!\left[f_0\in S\ \land\ \bigl(f_3\in S\ \lor\ f_4\in S\ \lor\ (f_2\in S\land f_1\notin S)\bigr)\right].
\label{eq:constructed-game}
\end{equation}
Feature $f_1$ has the harmful witness $S^*=\{f_0,f_2\}$ and is the only feature with exact minimum marginal $-1$. Algorithm A samples this witness with probability $p_-=1/[5\binom{4}{2}]=1/30$. Condition Algorithm B's perturbation kernel on the oracle seed coalition $S^*$ rather than on MI-correlation guidance. Its probability of returning exactly $S^*$ is
$\rho=.49/5+2[.21/(5\cdot3)]+.09/(5\cdot6)=.129$.
Figure~\ref{fig:overview}(b) therefore isolates the perturbation kernel; it is not real-data evidence that MI-correlation guidance is better. In the removal experiments, a hybrid dispatcher chooses the estimator from the analyzed dimension: exact enumeration for $n\leq20$, Algorithm B for $20<n\leq75$ (Algorithm A if no correlation neighbors are available), and the featurewise minimum of Algorithms A and B for $n>75$. Taking the minimum keeps every observed $-1$ witness-certified. On the six datasets this means exact enumeration on Abalone, Algorithm B on Spambase and Credit-Card-Fraud, and the A--B minimum on Madelon, ISOLET, and Gisette.

\subsection{Quota-Indexed Selection and Validation Gate}
\label{sec:selection}

For each quota, the estimated harmful set $\widehat{\mathcal{H}}_\alpha=\{i:\widehat{\operatorname{MinPI}}_i=-1\}$ proposes one compression--utility operating point. Sweeping $\alpha\in\{.85,.90,.95,.99\}$ traces several such points, which need not be monotone in $\alpha$. The removal rule uses the prespecified default $\alpha=.95$.

Every coalition model is fitted on the training split and its game utility is evaluated on validation data. The same validation split is therefore reused adaptively to construct the harmful set, apply the removal gate, and drive multi-objective search with the non-dominated sorting genetic algorithm NSGA-II~\citep{deb2002fast}. Because the quota, the gate margins, and the backoff rule are fixed in advance, validation applies these decisions rather than tuning them. The held-out split is reserved for final evaluation and is never consulted during selection. The candidate subset may trail the full model on validation by at most $.005$ for classification and $.01$ for regression. When the raw set fails, the all-or-none backoff removes none of its candidates. The accepted count also defines the deletion budget for SHAP-bottom, MinShap-bottom, Banzhaf-bottom, and random-matched controls. Appendix~\ref{app:algorithms} gives pseudocode for both estimators, the hybrid dispatcher, the gate, and the controls.

\section{Experimental Design}
\label{sec:experiments}

\subsection{Data, Models, and Scope}

Table~\ref{tab:datasets} summarizes the six retained benchmark datasets. The original campaign included seven datasets; following a post-hoc integrity audit, we excluded Communities \& Crime from the publication analysis because its target is reconstructible from retained outcome-component columns. All reported aggregates were recomputed from canonical records for the six retained datasets. Each seed uses a 64/16/20 train/validation/test split, stratified for classification. Variance filtering, mutual-information selection to at most 100 features, and standardization are fitted on training data only. The frozen seeds are $42,123,456,789,1024$.

\begin{table}[H]
\caption{Datasets and configured dimension cap. The realized analyzed dimension can be smaller after training-only variance filtering. OVR: one-vs-rest.}
\label{tab:datasets}
\centering
\footnotesize
\begin{tabular}{lrrrl}
\toprule
Dataset & Native dim. & Cap & Samples & Task / utility \\
\midrule
Spambase & 57 & 57 & 4,601 & binary / ROC AUC \\
Madelon & 500 & 100 & 2,600 & binary / ROC AUC \\
ISOLET & 617 & 100 & 7,797 & 26-class / OVR AUC \\
Gisette & 5,000 & 100 & 7,000 & binary / ROC AUC \\
Credit-Card-Fraud & 29 & 29 & 284,807 & binary / ROC AUC \\
Abalone & 8 & 8 & 4,177 & regression / $R^2$ \\
\bottomrule
\end{tabular}
\end{table}

All tasks use XGBoost with probabilities for classification, 100 boosting rounds, depth 4, learning rate $.1$, one CPU worker per fit, and the CUDA tree backend. The 100-round choice was selected from $\{10,20,40,100\}$ using training/validation-only calibration before formal test evaluation. Binary datasets use ROC AUC, ISOLET uses one-vs-rest AUC, and Abalone uses $R^2$; ISOLET accuracy appears only in a seed-42 utility ablation.

The reported removal and full-versus-prefiltered NSGA-II analyses cover all six retained benchmark datasets and five seeds. Both NSGA-II~\citep{deb2002fast,hamdani2007multi} variants minimize $|S|/$$n$ and $1-\operatorname{clip}(U,0,1)$, where $n$ is the realized analyzed dimension before optional prefiltering and is shared by both variants. They use population 40, 800 evaluations, and reference point $(1.1,1.1)$. Estimator budgets are $\{2,5,10,25\}$ sampled coalitions (Algorithm A) or perturbations (Algorithm B) per feature, so $K=K_{\mathrm{adv}}$ equals the budget, except ISOLET, which omits 25; removal uses the hybrid dispatcher at the largest budget; exact enumeration is restricted to at most 16 features and therefore provides formal ground truth only on Abalone. The $\alpha$ sweep covers all six retained datasets at seed 42. Guidance and dispatcher ablations cover Madelon, ISOLET, Credit-Card-Fraud, and Abalone at seed 42; the utility ablation is restricted to ISOLET at seed 42.

\subsection{Comparisons, Statistics, and Provenance}

The removal comparison includes full, MinPI, SHAP-bottom~\citep{lundberg2017unified}, MinShap-bottom~\citep{zheng2026minshap}, Banzhaf-bottom~\citep{kulynych2017feature}, and random-matched subsets. SHAP-bottom deletes the $k$ features with the smallest mean absolute TreeSHAP value of the full XGBoost model on the training split; MinShap-bottom and Banzhaf-bottom delete the $k$ features with the smallest minimum and mean utility marginal, respectively, estimated from $K$ shared sampled coalitions. For a retained subset, $\Delta U$ denotes its held-out utility minus that of the full model, so positive values favor removal. Thirty random-matched subset trials are averaged within seed, preserving five paired seed-level replicates; each trial draws $k$ features uniformly without replacement, where $k$ is the accepted MinPI-matched removal count, deterministically seeded by the dataset--seed unit and trial index. Dataset-level tables report means and sample standard deviations. Across datasets, removal ranks use each method's five-seed dataset mean; a Friedman test and paired Wilcoxon tests with Holm adjustment are descriptive because there are only six blocks.

NSGA-II endpoints are validation hypervolume (HV), held-out hypervolume of the validation-selected front, and a post-hoc attainment count: the first evaluation at which cumulative-best validation hypervolume reaches 95\% of that run's final-best value. Exact paired sign-flip tests align the five seed replicates; Holm correction is applied within each seven-comparison endpoint family (six datasets plus the dataset-averaged result). The original seven-dataset campaign completed every planned job in the five-seed core stages and the seed-42 sensitivity stages. Protocol and provenance details are given in the Reproducibility Details in the appendix.

\begin{figure}[htbp]
\centering
\includegraphics[width=\linewidth]{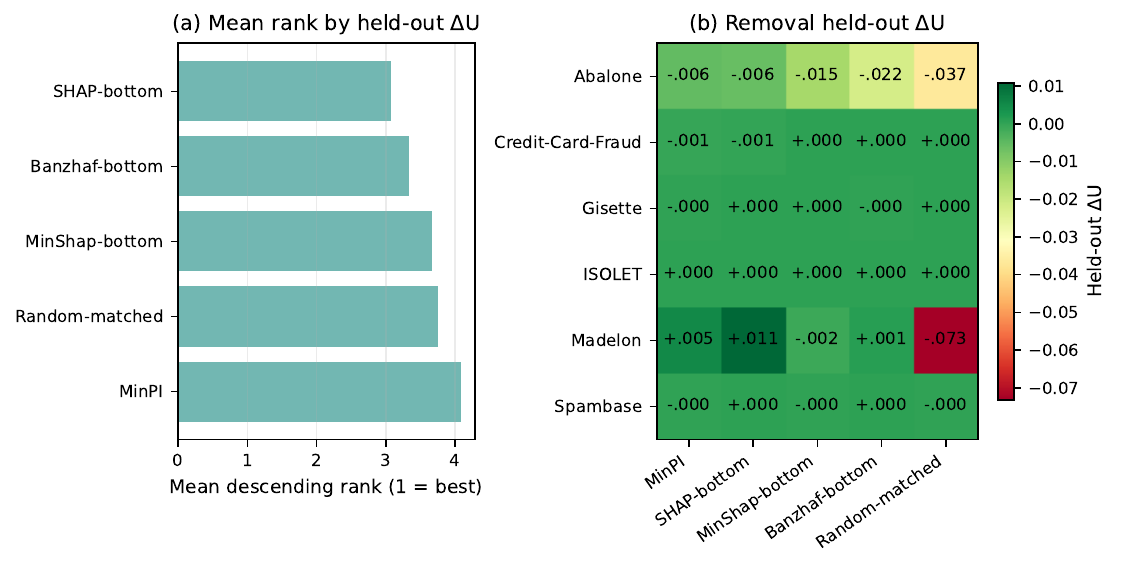}
\caption{Validation-gated matched removal. (a) Mean descending ranks (1 is best) of the six methods from their five-seed mean $\Delta U$ (full participates but is omitted). (b) Five-seed mean $\Delta U$; positive favors removal. Controls delete the accepted MinPI-matched count, with 30 random-matched trials averaged per seed. }
\label{fig:removal}
\end{figure}

\section{Results}
\label{sec:results}

\subsection{RQ1: Validation-Gated Removal}

Table~\ref{tab:removal} shows heterogeneous compression: averaged over the 30 dataset--seed units, MinPI removed 12.53\% of analyzed features at a negligible mean held-out utility cost of $-.000313$ (nonzero removal in 14 of 30 units). Before the gate, MinPI flagged on average 48.0 of 100 Madelon features and 4.4 of 8 Abalone features, and at most two features in any seed of the other datasets (none on ISOLET). Madelon had the largest compression (43.6\%) with a positive $\Delta U=+.00469$, Abalone compressed by 27.5\% with $\Delta U=-.00565$, and Credit-Card-Fraud by \CreditCompression{} with $\Delta U=\CreditDeltaUtility{} $; relative to the same-budget random-matched control, MinPI's removals were markedly preferable on the two most-compressed datasets ($+.0047$ versus $-.0732$ on Madelon, $-.0056$ versus $-.0370$ on Abalone). The raw harmful set passed the validation gate in 27 of 30 units (13 with no candidates, 14 of 17 nonempty passed); the failed cases backed off to zero removal, showing the gate is substantive.

\begin{table}[htbp]
\caption{MinPI removal at $\alpha=.95$ over five seeds. Compression and held-out $\Delta U$ are mean $\pm$ SD; ``raw pass'' counts validation-gate passes before backoff. Harmful and removed are mean counts of features flagged before the gate and removed after it.}
\label{tab:removal}
\centering
\footnotesize
\begin{tabular}{lrrrrc}
\toprule
Dataset & Harmful & Removed & Compression (\%) & Held-out $\Delta U$ & Raw pass \\
\midrule
Abalone & 4.4 & 2.2 & $27.50\pm25.62$ & $-.00565\pm.01077$ & $3/5$ \\
Credit-Card-Fraud & 0.8 & 0.8 & $2.76\pm2.89$ & $-.00078\pm.00335$ & $5/5$ \\
Gisette & 0.6 & 0.6 & $.60\pm.89$ & $-.00011\pm.00016$ & $5/5$ \\
ISOLET & 0.0 & 0.0 & $.00\pm.00$ & $.00000\pm.00000$ & $5/5$ \\
Madelon & 48.0 & 43.6 & $43.60\pm31.96$ & $+.00469\pm.00497$ & $4/5$ \\
Spambase & 0.4 & 0.4 & $.72\pm.99$ & $-.00003\pm.00007$ & $5/5$ \\
\bottomrule
\end{tabular}
\end{table}

\begin{table}[t]
\caption{Five-seed mean held-out utility after each variant deletes the accepted MinPI-matched count; utility is AUC/OVR-AUC/$R^2$ by task, higher is better. SHAP/MinShap/Banzhaf/Random/Credit abbreviate SHAP-bottom/MinShap-bottom/Banzhaf-bottom/random-matched/Credit-Card-Fraud.}
\label{tab:utility}
\centering
\footnotesize
\begin{tabular}{lrrrrrr}
\toprule
Dataset & Full & MinPI & SHAP & MinShap & Banzhaf & Random \\
\midrule
Madelon & .7876 & .7923 & .7985 & .7861 & .7884 & .7144 \\
Abalone & .5397 & .5340 & .5333 & .5251 & .5174 & .5027 \\
Credit & .9734 & .9726 & .9728 & .9736 & .9736 & .9736 \\
Gisette & .9887 & .9886 & .9887 & .9889 & .9887 & .9887 \\
ISOLET & .9935 & .9935 & .9935 & .9935 & .9935 & .9935 \\
Spambase & .9858 & .9858 & .9858 & .9856 & .9859 & .9858 \\
\bottomrule
\end{tabular}
\end{table}

Table~\ref{tab:utility} lists the held-out utilities achieved by the removal variants as a baseline comparison on the same deletion budget; Figure~\ref{fig:removal} shows the resulting mean ranks and per-dataset $\Delta U$.

MinPI's average utility across datasets was essentially identical to the full model (descriptive difference $-0.0003$, Wilcoxon $p = 0.44$). Its mean rank among all methods was 4.08, and neither the Friedman test ($p = 0.89$) nor the pairwise Wilcoxon tests of MinPI against each other method (raw $p \geq 0.4375$, Holm-adjusted $p = 1.0$) indicated a significant method effect.

Removal thus cost almost no held-out utility on average, but it did not improve on the full model. The mean rank of 4.08 reflects the ordinal scale: MinPI ranks last on Credit-Card-Fraud and Gisette and fifth on Spambase, where every method lies within $.001$ of the full model, but second on Madelon and Abalone, where it beats random-matched removal by $.078$ and $.031$.

Across datasets, the method's value scales with the amount of removable structure: on Madelon, whose 500 features are dominated by irrelevant variables, it removed $43.6\%$ with positive held-out $\Delta U$; on ISOLET it removed nothing; on Credit-Card-Fraud, Gisette, and Spambase removals were small and near-lossless; and on Abalone, a small noisy-target regression, its $27.5\%$ removal slightly reduced utility, suggesting weakly informative features can be misclassified as harmful. The validation gate is a substantive safeguard, vetoing the raw harmful set in the two Abalone seeds and one Madelon seed that failed the loss bound. Compression is seed-sensitive (SDs of $25$--$32$ percentage points on Madelon and Abalone), so the harmful set should be read as a five-seed aggregate.

\subsection{RQ2: Quota Sensitivity}

\begin{figure}[t]
\centering
\includegraphics[width=\linewidth]{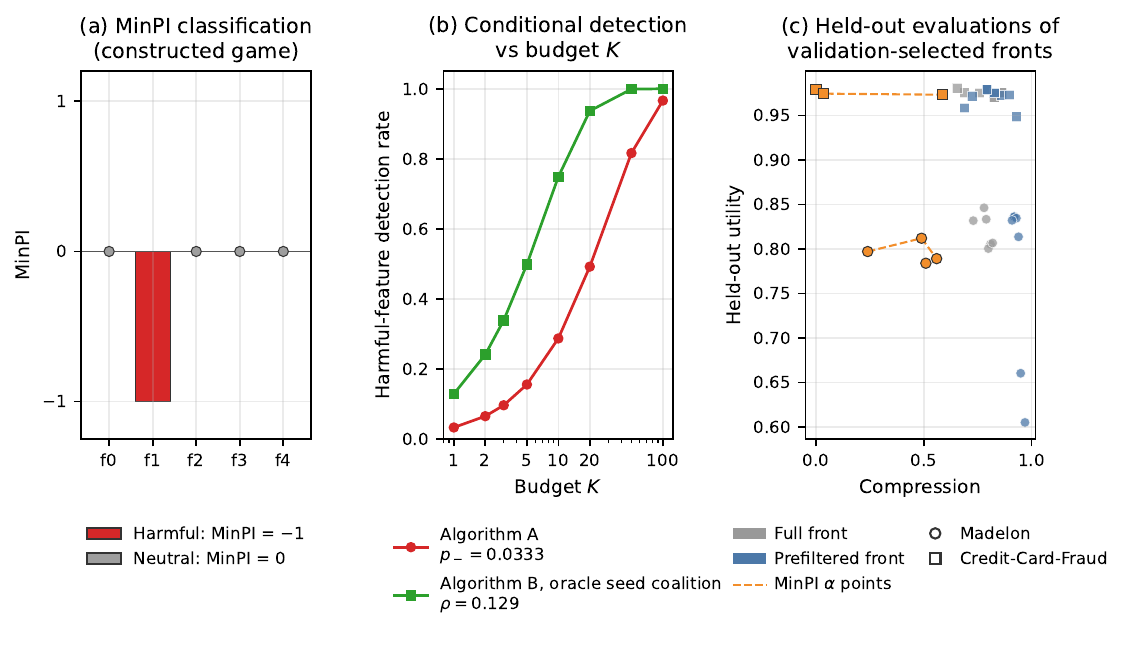}
\caption{Overview of MinPI. (a) Exact minimum marginals in the constructed game~\eqref{eq:constructed-game}: $f_1=-1$, the rest neutral. (b) Detection probability vs budget for Algorithm A ($p_-=1/30=.0333$) and Algorithm B on the oracle seed coalition $\{f_0,f_2\}$ ($\rho=.129$); illustrative, not real-data guidance evidence. (c) Seed-42 held-out fronts of full vs prefiltered NSGA-II with the $\alpha$ points on Madelon and Credit-Card-Fraud; connecting lines only order $\alpha$.}
\label{fig:overview}
\end{figure}

The $\alpha$ sweep is a single-seed (seed 42) sensitivity analysis. Its trajectories were dataset-dependent and non-monotone. On Madelon, compression across $\alpha=(.85,.90,.95,.99)$ was $(.24,.49,.56,.51)$ with held-out $\Delta U=(+.0077,+.0227,-.0003,-.0054)$. On Abalone, compression was $(.375,.50,.50,0)$, while on Credit-Card-Fraud it rose from 0 at $.85$ and $.90$ to $.034$ at $.95$ and $.586$ at $.99$. Figure~\ref{fig:overview}(c) overlays the points for Madelon and Credit-Card-Fraud on held-out evaluations of their seed-42 validation-selected fronts; the complete six-dataset sweep is reported in Table~\ref{tab:alpha} in the appendix.

\subsection{RQ3: Prefiltered Multi-Objective Search}

With the same 800-evaluation budget, MinPI-prefiltered NSGA-II changed the dataset-averaged held-out hypervolume by $+.01584$ and validation hypervolume by $+.01741$. For each endpoint, all five seed-level dataset-averaged differences were positive (rank-biserial correlation 1), but the exact sign-flip $p=.0625$ became $.4375$ after the Holm correction. Dataset mean held-out HV differences were $+.09515$ on Madelon, $+.00535$ on Credit-Card-Fraud, $+.00021$ on Abalone, 0 on ISOLET, $-.00191$ on Gisette, and $-.00378$ on Spambase. Thus the positive dataset-averaged direction is driven mainly by Madelon, with only small positive differences on Credit-Card-Fraud and Abalone.

\begin{figure}[t]
\centering
\includegraphics[width=\linewidth]{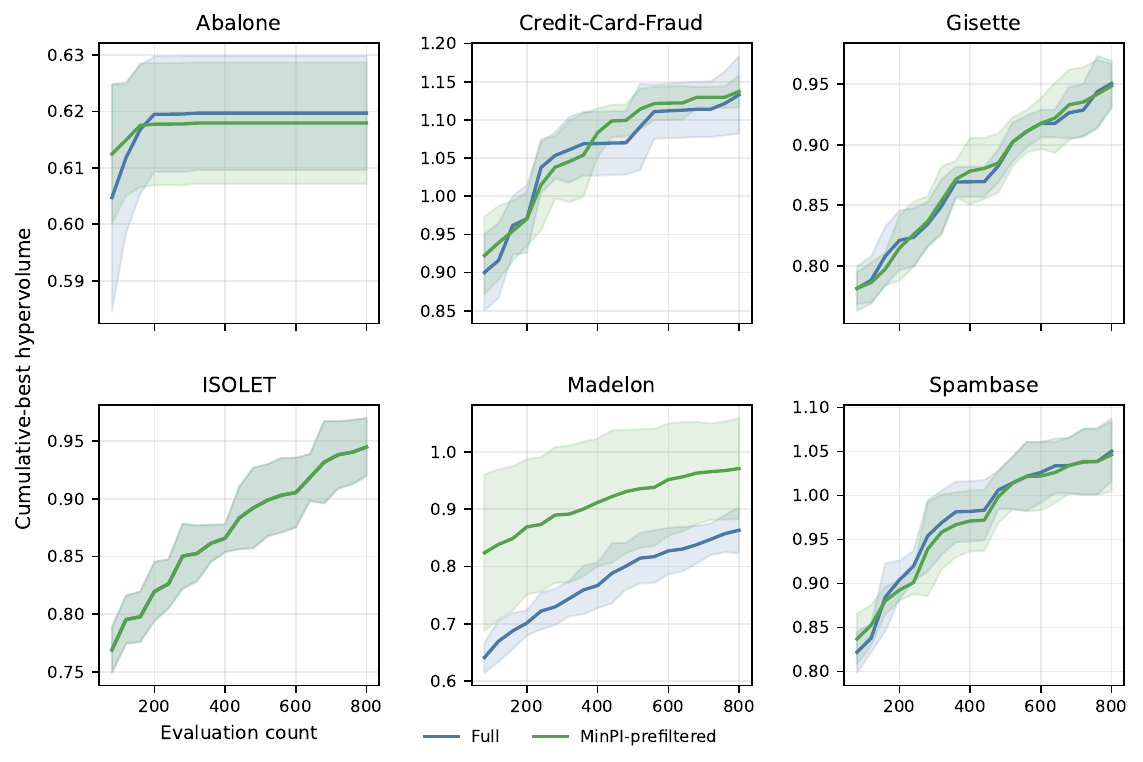}
\caption{Hypervolume convergence of full vs MinPI-prefiltered NSGA-II under the matched 800-evaluation budget. Each panel is a dataset; blue/green lines are five-seed means of cumulative-best validation hypervolume with $\pm$1-SD shading and reference point $(1.1,1.1)$. Curves coincide on most datasets and separate on Madelon; the dataset-averaged held-out change is \NsgaMeanHvChange{}. No dataset-wise comparison is significant after Holm correction.}
\label{fig:prefilter}
\end{figure}

On average, the post-hoc diagnostic reached 95\% of the final HV 22.67 evaluations earlier under MinPI prefiltering (Figure~\ref{fig:prefilter}), with the largest reductions on Madelon and Gisette and a slight increase on Spambase.
Both variants used the same total budget of 800 evaluations. Prefiltering thus improved search on Madelon, where it removed the most features, and changed it little elsewhere.

\subsection{RQ4: Estimator, Guidance, and Utility Sensitivity}

Abalone is the only dataset small enough for exact enumeration. Averaged over the five seed-level F1 scores of the estimated harmful set against the exact one, at budgets $2,5,10,25$, Algorithm A achieved $.100,.150,.230,.443$, while Algorithm B (MI-correlation guidance) achieved $.000,.080,.133,.313$. The hybrid dispatcher scored 1 at every budget because it selected exact enumeration at eight features. Thus the five-seed mean F1 for Algorithm B was lower than for Algorithm A at every finite budget.

The ablations likewise show sensitivity rather than a stable guidance ordering. On Madelon, random, MI-only, correlation-only, and MI-correlation seed coalitions recovered 5, 16, 34, and 28 harmful labels; on Credit-Card-Fraud the counts were 6, 2, 2, and 1. On ISOLET, changing the utility while keeping the probability-producing classifier fixed changed the harmful count from 0 under the formal OVR AUC to 9 under accuracy. These single-seed results show that estimator and utility choices materially affect recovered labels.

 Exact enumeration and the hybrid dispatcher recover every harmful label on Abalone, so exact enumeration should be used whenever the dimension allows it. Beyond that, Algorithm A is the safer default: it reaches every coalition, whereas Algorithm B did not improve on it where this could be checked. The sensitivity of the recovered harmful set to estimator, seed construction, and utility likewise reinforces that MinPI should be consumed as a conservative, validation-gated, multi-seed aggregate rather than a single configuration.

\FloatBarrier

\section{Discussion}
\label{sec:discussion}

MinPI's main contribution is its semantics: an exact or sampled $-1$ is witness-certified by an evaluated threshold-crossing coalition, the validation gate separates witness discovery from removal, and finite sampling preserves witness validity but not completeness (a sampled 0 or $+1$ does not certify the exact class). MinPI is a worst-case support statistic rather than an interaction index~\citep{grabisch1999axiomatic,rabitti2019shapley}, certifying a single harmful witness per feature rather than quantifying pairwise interactions.

In prefiltered search: Madelon drives most of the positive average hypervolume change, Credit-Card-Fraud and Abalone show small positive differences, and Gisette and Spambase small decreases. Prefiltering is useful when its accepted set removes distracting dimensions.

These findings delimit where MinPI is most useful. It removes only witness-certified harmful features under a gate that vetoes any removal failing a fixed loss bound, concentrating its strongest results on datasets with abundant removable structure, removing little or nothing elsewhere, and benefiting most from the gate on small noisy-target problems. The natural applications are (i) prefiltering high-dimensional feature spaces before modeling or multi-objective search, (ii) auditing feature quality to flag suspicious, noisy, or leak-prone predictors, and (iii) a validation-gated removal step for single-task models. MinPI does not require perfectly clean data---it requires data in which removable structure exists---and its output should be consumed as a five-seed aggregate subject to the gate.

Several limitations constrain interpretation. First, five paired seeds and six datasets give low-powered tests, and the ablations use only seed 42. Second, a single XGBoost family and mixed utility scales make a common multiplicative quota non-invariant to utility shifts. Third, exact estimator comparison is available only on eight-feature Abalone, where Algorithm B's F1 was below Algorithm A's at every budget; the detection bound~\eqref{eq:detection} is a support condition, not evidence that guidance assigns more mass to harmful witnesses. Fourth, shared caching complicates runtime attribution, so we make no speedup claim. Fifth, the validation gate bounds validation loss only and does not guarantee held-out non-inferiority. Finally, paired seed-level variation, broader model families, a stronger guidance estimator, standard filter/wrapper baselines, and extending harmful-feature removal to multi-task settings remain future work.

\section{Conclusion}

Exact MinPI specializes minimum marginal aggregation to a binary utility-threshold game and yields a quota-conditioned three-way classification; finite-budget estimators certify only observed harmful threshold crossings. The resulting candidates feed a validation-gated removal or prefiltering step. Across the six benchmark datasets, the method removed 12.5\% of features on average (43.6\% on Madelon) with a mean held-out utility change of $-0.0003$, lost far less utility than random removal of the same size on Madelon and Abalone, and yielded a positive although non-significant average hypervolume change in prefiltered search. MinPI is a conservative feature-deletion strategy whose clearest gains are concentrated on datasets with abundant removable structure, and whose value otherwise depends on the model, split, dataset, estimator, utility, quota, and validation decision---consistent with the observation that harmful features are context-dependent rather than universally removable~\citep{lippl2024inductive}.

{
\section*{AI Disclosure Statement}
The authors used Claude Code and ChatGPT for language refinement and code assistance. The authors reviewed and verified all text, references, code and figures, and take full responsibility for the content of this paper.
}

\bibliography{icomp2026_conference}
\bibliographystyle{icomp2026_conference}

\appendix

\section{Algorithms}
\label{app:algorithms}

{Algorithms~\ref{alg:a}--\ref{alg:removal} restate the estimators, the hybrid dispatcher, and the validation-gated removal step as implemented. $\Delta_i(S)$ is the marginal of $v_\alpha$ in~\eqref{eq:utvg}, whose coalition models are fitted on the training split and evaluated on the validation split.}

\begin{algorithm}[htbp]
\caption{Size-uniform coalition sampling for feature $i$ (Algorithm A)}
\label{alg:a}
\begin{algorithmic}[1]
\Require analyzed features $N$, feature $i$, budget $K$
\State $m\gets+1$
\For{$t=1,\ldots,K$}
\State draw $s$ uniformly from $\{0,\ldots,n-1\}$
\State draw $S$ uniformly from the $s$-element subsets of $N\setminus\{i\}$
\State $m\gets\min\{m,\Delta_i(S)\}$
\If{$m=-1$} \Return $-1$ with witness $S$ \EndIf
\EndFor
\State \Return $m$
\end{algorithmic}
\end{algorithm}

\begin{algorithm}[htbp]
\caption{MI-correlation-guided perturbation for feature $i$ (Algorithm B)}
\label{alg:b}
\begin{algorithmic}[1]
\Require $N$, feature $i$, budget $K_{\mathrm{adv}}$, training MI scores, maximum-weight correlation tree on the training features
\State $B_i\gets$ up to ten tree neighbors of $i$, by decreasing absolute correlation
\State add features of the top-MI half of $N\setminus\{i\}$, by decreasing MI, until $|B_i|\geq\min(20,\lfloor n/3\rfloor)$
\State $m\gets+1$
\For{$t=1,\ldots,K_{\mathrm{adv}}$}
\State $S\gets B_i$ with each member removed independently with probability $.3$
\State draw $a$ uniformly from $\{0,\ldots,\min(5,n-1)\}$
\State add to $S$ the first $a$ features of a random order of $N\setminus\{i\}$ that are not in $S$
\State $m\gets\min\{m,\Delta_i(S)\}$
\If{$m=-1$} \Return $-1$ with witness $S$ \EndIf
\EndFor
\State \Return $m$
\end{algorithmic}
\end{algorithm}

\begin{algorithm}[htbp]
\caption{Hybrid dispatcher, validation gate, and matched controls}
\label{alg:removal}
\begin{algorithmic}[1]
\Require training and validation splits, quota $\alpha$, budget $K$, margin $\varepsilon$ ($.005$ for classification, $.01$ for regression)
\If{$n\leq20$}
\State compute $\operatorname{MinPI}_i$ for every $i$ by exact enumeration
\ElsIf{$n\leq75$}
\State run Algorithm B for every $i$ (Algorithm A if no correlation tree is available)
\Else
\State run Algorithms A and B for every $i$ and take the featurewise minimum
\EndIf
\State $H\gets\{i:\widehat{\operatorname{MinPI}}_i=-1\}$
\State fit the model on $N\setminus H$ and evaluate its validation utility $U_{\mathrm{val}}(N\setminus H)$
\If{$U_{\mathrm{val}}(N\setminus H)\geq U_{\mathrm{val}}(N)-\varepsilon$}
\State $R\gets H$
\Else
\State $R\gets\emptyset$ \Comment{all-or-none backoff}
\EndIf
\State $k\gets|R|$
\State SHAP-, MinShap- and Banzhaf-bottom delete the $k$ lowest-scoring features; random-matched deletes $k$ features drawn uniformly, in each of 30 trials
\State \Return $N\setminus R$, evaluated once on the held-out split
\end{algorithmic}
\end{algorithm}

\FloatBarrier
\section{Quota Sweep}

Table~\ref{tab:alpha} reports the seed-42 $\alpha$ sweep (compression, held-out $\Delta U$) for the six benchmark datasets; the non-monotone rows complement the RQ2 discussion.

\begin{table}[H]
\caption{Quota operating points, shown as (compression, held-out $\Delta U$). Every cell has $n=1$.}
\label{tab:alpha}
\centering
\scriptsize
\begin{tabular}{lcccc}
\toprule
Dataset & $\alpha=.85$ & $\alpha=.90$ & $\alpha=.95$ & $\alpha=.99$ \\
\midrule
Abalone & $(.375,+.0041)$ & $(.500,-.0198)$ & $(.500,+.0010)$ & $(0,0)$ \\
Credit-Card-Fraud & $(0,0)$ & $(0,0)$ & $(.034,-.0045)$ & $(.586,-.0057)$ \\
Gisette & $(0,0)$ & $(0,0)$ & $(0,0)$ & $(.090,+.0002)$ \\
ISOLET & $(0,0)$ & $(0,0)$ & $(0,0)$ & $(.010,-.000005)$ \\
Madelon & $(.240,+.0077)$ & $(.490,+.0227)$ & $(.560,-.0003)$ & $(.510,-.0054)$ \\
Spambase & $(0,0)$ & $(0,0)$ & $(0,0)$ & $(.196,-.0012)$ \\
\bottomrule
\end{tabular}
\end{table}

\section{Reproducibility Details}

The formal splits are 64\%/16\%/20\% train/validation/test; dataset identities, split indices, transformations, model policy, source, and scientific environment are recorded with cryptographic digests, and resume rejects protocol drift. Training data fit preprocessing, guidance, and all coalition models; validation is reused adaptively for threshold-game construction, the removal gate, and NSGA-II selection, while the held-out partition is read only for final evaluation. The original seven-dataset campaign is retained as the historical record; a post-hoc integrity audit excluded Communities \& Crime from the publication analysis for outcome-component leakage, and the reported tables and figures were recomputed for the six retained datasets. The estimator set, removal controls, and NSGA-II variants are as in Section~3; runtime counters are diagnostic only. Tables and figures are rederived from canonical row-level records and independently hash-verified.

\end{document}